\documentclass[conference]{IEEEtran}
\IEEEoverridecommandlockouts
\usepackage{cite}
\usepackage{amsmath,amssymb,amsfonts}
\usepackage{algorithmic}
\usepackage{graphicx}
\usepackage{textcomp}
\usepackage{xcolor}
\usepackage{booktabs}
\usepackage{multirow}
\usepackage{url}
\usepackage{threeparttable}
\usepackage{hyperref}
\usepackage{float}
\usepackage{subcaption}
\def\BibTeX{{\rm B\kern-.05em{\sc i\kern-.025em b}\kern-.08em
    T\kern-.1667em\lower.7ex\hbox{E}\kern-.125emX}}
\begin{document}
\title{Performance Anomaly Detection in Athletics: \\ A Benchmarking System with Visual Analytics}

\author{\IEEEauthorblockN{Blessed Madukoma}
\IEEEauthorblockA{\textit{Carnegie Mellon University Africa}\\
Kigali, Rwanda \\
bmadukom@andrew.cmu.edu}
\and
\IEEEauthorblockN{Prasenjit Mitra}
\IEEEauthorblockA{\textit{Carnegie Mellon University Africa}\\
Kigali, Rwanda \\
prasenjitm@andrew.cmu.edu}
}

\maketitle
\begin{abstract}
Anti-doping programs rely on biological testing to detect performance-enhancing
drugs, but such testing costs over \$800 per sample and is limited by short detection windows for many prohibited substances. These constraints leave
large portions of athletes without regular testing, motivating complementary
screening approaches that analyze routine competition results to identify
suspicious performance patterns. We present a system that processes 1.6 million
athletics performances from 19,000+ competitions (2010--2025) using eight detection
methods ranging from statistical rules to machine learning and trajectory
analysis. We validate all methods against publicly confirmed anti-doping
violations to measure their effectiveness in identifying sanctioned athletes.
Trajectory-based methods, which compare performances to expected career
progression, achieve the best balance between detecting violations and limiting
false alarms, though all methods face challenges from incomplete data and rare
confirmed violations. The system provides an interactive interface for
expert-driven investigation, emphasizing transparency and human judgment to
support, rather than replace, established anti-doping processes.
\end{abstract}

\section{Introduction}
Competitive sports face ongoing challenges from performance-enhancing drugs, which give unfair advantages and undermine fair play \cite{wada_isti_2023,hopker2018performance}. Anti-doping programs aim to detect and deter violations through biological testing \cite{wada_isti_2023}, but tests cost over \$800 per sample \cite{asoif_costs_2010,maennig2014} and some substances (e.g., testosterone, EPO) are detectable for only hours or days after administration \cite{thevis2013_banned_review}. Because testing is expensive and detection windows are short, programs focus on high-risk athletes and priority competitions, leaving much of the population untested \cite{wada_isti_2023,overbye2016_doping_control}. This motivates complementary approaches that screen routine competition results continuously \cite{iljukov2017_performance_profiling,hopker2020_performance_profiling,ryoo2022_weightlifting}.

This work tackles the athlete screening problem: given millions of performances across thousands of competitions, identify athletes who merit closer anti-doping review. We define the task as flagging performances that deviate from an athlete’s expected trajectory based on times, dates, and environmental conditions. Performance is evaluated using athlete-level precision (fraction of flagged athletes with confirmed sanctions), recall (fraction of sanctioned athletes detected), and F1 score. For context, biological testing has a low positive rate of 0.80\%  \cite{wada_testing_figures_2023}, so performance-based screening with similar detection rates complements existing programs. Because false accusations can harm athletes’ reputations, sponsorships, and careers \cite{hard2009_athletes_rights,woolway2020_legitimacy}, detection methods should support cautious, expert-led review rather than automated enforcement \cite{wada_isti_2023}.

\begin{figure}[!htb]
\centering
\includegraphics[width=\columnwidth]{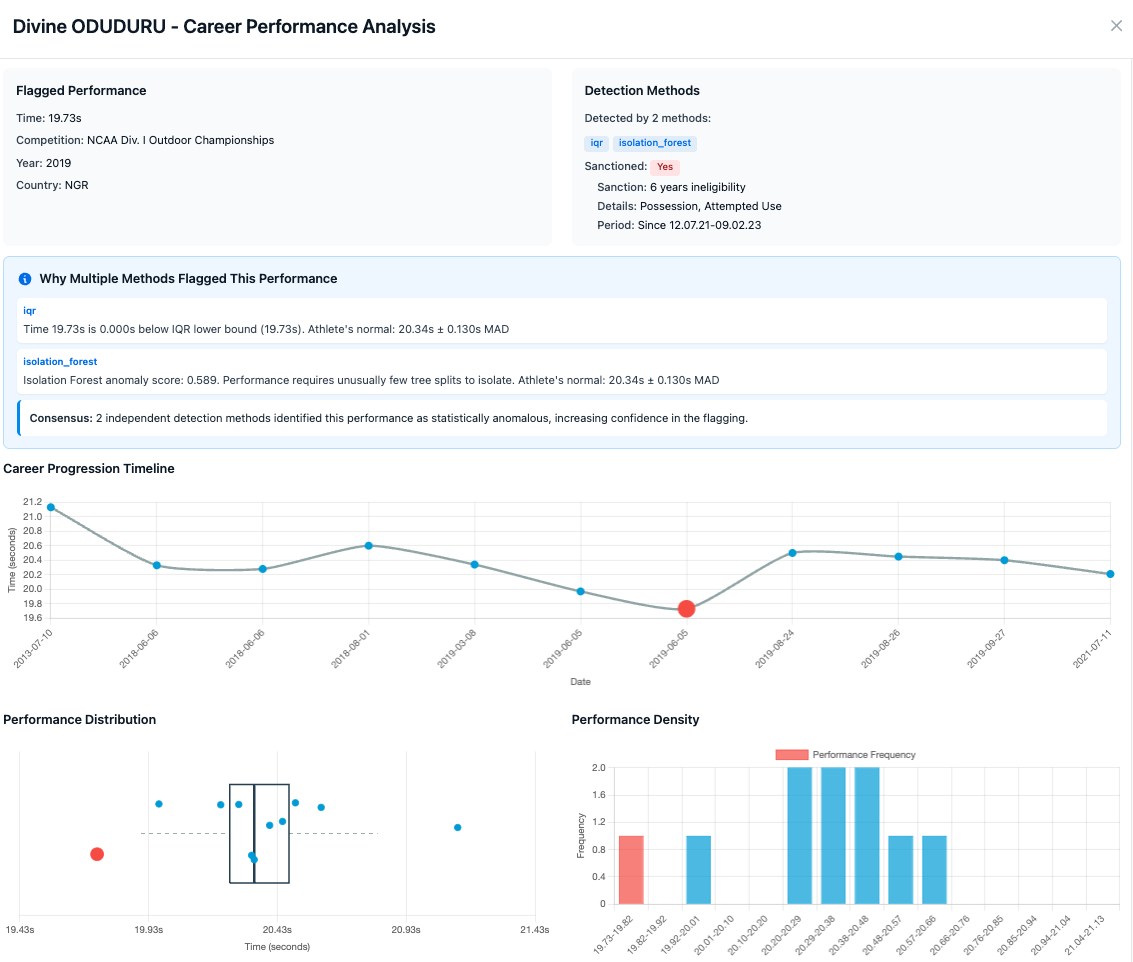}
\caption{Multi-method consensus flagging of a sanctioned 200\,m sprinter's 19.73\,s performance (2019 NCAA Championships) by IQR Method and Isolation Forest (score 0.589; normal: 20.34\,s $\pm$ 0.130\,s MAD). The interface shows method explanations, career timeline, and distributions. A later 6-year sanction confirms the flag. This transparency helps investigators assess cross-method agreement before prioritizing testing.}
\label{fig:teaser}
\end{figure}

Monitoring the performance of athletes offers several advantages for this screening problem. First, it directly measures race times, the outcome that anti-doping programs seek to protect, as the primary goal of doping is to improve competitive performance
\cite{montagna2018_bayesian,hopker2024_excess_performance}. Secondly,
performance data are collected routinely at all competition levels, enabling
continuous screening without additional testing infrastructure. Finally,
performance signals can help prioritize limited testing resources by
identifying athletes whose performance patterns warrant closer examination
\cite{wada_isti_2023,iljukov2017_performance_profiling}. When combined with
the Athlete Biological Passport (which monitors blood biomarkers longitudinally
\cite{wada_abp_og_2023}), performance signals can contribute to targeted testing
and investigative case review \cite{wada_abp_og_2023,
worldathletics_abp_notes_2014}. To support this investigative approach, our
system provides transparent multi-method detection with expert review
capabilities: when a performance is flagged, investigators see method-specific
explanations and contextual competition details rather than opaque automated
scores, as shown in Figure~\ref{fig:teaser}.

The core technical challenge in performance-based screening is that athletic performance varies naturally
for many legitimate reasons, and environmental factors can systematically
shift sprint outcomes. For example, altitude provides measurable advantages in the 100\,m. Empirical analysis of Olympic Games results indicates time improvements of
approximately 0.19 seconds at 2250~m elevation for men and 0.21 seconds for
women \cite{linthorne2016_altitude}. Additionally, wind assistance affects performance, with
analyses estimating time advantages of approximately 0.10 seconds for men and
0.12 seconds for women at the maximum legal tailwind of +2.0~m/s
\cite{linthorne1994_wind}. Competition round also matters, as athletes may
not perform maximally in preliminary heats \cite{hanley2021_mesopacing}.
Simple statistical rules (e.g., flagging performances more than 3 standard
deviations from an athlete's mean) ignore athlete-specific variance and fail
to account for context (wind, altitude, round), generating systematic false
positives \cite{hoaglin1983robust,hastie2009elements}.

Despite over a decade of research on performance profiling for anti-doping
\cite{schumacher2009profiling}, systematic cross-method comparison remains absent, evaluation protocols lack standardization, and interactive decision-support tools are underdeveloped. First, most
studies evaluate only 1 to 3 detection methods on a single sport, making
systematic comparison difficult
\cite{iljukov2017_performance_profiling,hopker2020_performance_profiling,
montagna2018_bayesian,ryoo2022_weightlifting}. Secondly, performance monitoring
is often treated as an investigative signal rather than as production-ready
decision support with interactive tools and reproducible evaluation protocols.
Finally, evaluation is complicated by incomplete ground truth: publicly
available sanctions reflect only detected violations and can arrive 2 to 8
years after the original performance
\cite{wada_sample_storage_2020,ita_london2012_reanalysis,
worldathletics_2015_retests}.
Among existing approaches, Bayesian hierarchical models address the challenge that athletes have vastly different amounts of competition data by simultaneously estimating
population-level performance distributions and individual athlete deviations,
borrowing statistical strength from the broader athlete population
\cite{montagna2018_bayesian,gelman2013bayesian}. However, these models
require repeated likelihood evaluations across thousands of MCMC sampling
iterations per athlete, making them computationally expensive for interactive
screening of tens of thousands of athletes. In contrast, the excess performance approach
measures an athlete's deviation from their expected career progression curve
\cite{hopker2024_excess_performance} and serves as our primary reproducible
published baseline due to its balance of detection quality and computational efficiency.

We present a performance anomaly detection system for athletics that integrates
large-scale data infrastructure, multiple detection methods, and visual
analytics to address these limitations. The data pipeline processes 1.6 million performance records from
19,000+ competitions (2010--2025), linking athlete identities and
integrating publicly available sanction records. The system implements eight
detection methods spanning statistical outlier rules (z-score, MAD, IQR),
machine learning algorithms (Isolation Forest, XGBoost),
trajectory-based temporal models (excess performance), and Bayesian
hierarchical inference. An interactive visual analytics interface supports
investigator-driven screening through side-by-side method comparison, career
trajectory visualization, and consensus-based case review.
Our quantitative evaluation focuses on 100\,m sprint performances while
demonstrating system generality through cross-event analysis on 200\,m and 400\,m sprints (Section~\ref{evaluation}). We benchmark all methods on the
same athlete population, ground truth labels, and evaluation protocol.

Our contributions are as follows:
\begin{itemize}
\item \textbf{A reproducible athletics data pipeline and benchmark.}
We provide a reproducible data pipeline that: (1)~links athlete identities
across 19,000+ competitions, (2)~supports event-specific performance slicing,
(3)~integrates publicly available sanction records as partial ground truth,
and (4)~enables reproducible benchmarking under identical data and protocol
conditions. This enables large-scale, repeatable evaluation not possible with
raw World Athletics results pages.

\item \textbf{A visual analytics system for anomaly screening and trajectory review.}
We provide a web-based interface for identifying and examining statistically
anomalous performances. The system supports screening (identifying performances
by event and time window that deviate from expected patterns) and trajectory
review (examining athlete performance histories with flagged anomalies).
Users can compare eight detection methods side-by-side, visualize performance
trajectories with method-specific explanations, and examine consensus views
showing cross-method agreement.

\item \textbf{Systematic benchmarking of reproducible detection methods.}
We provide the first systematic comparison of statistical, machine learning,
and Bayesian methods on the same large-scale dataset, evaluating eight detection
methods under identical conditions on 31,604 athletes for 100\,m sprints.
Methods are evaluated using athlete-level precision, recall, and F1 scores
against publicly confirmed sanctions. Excess performance
\cite{hopker2024_excess_performance} serves as the primary reproducible
trajectory-based baseline, and we include classical statistical rules
(z-score, MAD, IQR) to quantify the performance gap between context-free
thresholds and trajectory-aware approaches. Results are reported for 100\,m 
sprints (detailed benchmark), 200\,m and 400\,m sprints (generality demonstration).

\end{itemize}

\section{Background and Prior Work}
This work applies anomaly detection methods to longitudinal athletics
performance data for anti-doping screening, building on: (1) general anomaly
detection methods \cite{chandola2009survey,goldstein2016comparative}, and
(2) applied work on using competition results to identify suspicious
performance patterns in elite sport
\cite{schumacher2009profiling,iljukov2017_performance_profiling,
hopker2018performance,hopker2020_performance_profiling}.
\subsection{Anomaly Detection and Domain Challenges}
Anomaly detection identifies observations deviating substantially from
expected patterns \cite{chandola2009survey,goldstein2016comparative}. Methods
span multiple paradigms: statistical rules flag observations beyond fixed
thresholds (e.g., values exceeding $3\sigma$ from the mean); machine learning
approaches score novelty using decision boundaries or density estimates
(isolation forests, local outlier factor); and probabilistic models quantify
deviation via likelihood or posterior predictive probabilities. These methods
have been applied across domains including fraud detection, industrial
monitoring, and sports performance analysis \cite{chandola2009survey}.
However, sports performance presents distinct challenges: outcomes vary
legitimately due to training, maturation, and environmental context,
requiring domain-aware adaptations rather than direct application of generic
algorithms.

\subsection{Performance Monitoring Approaches}
Within sports integrity, performance monitoring tracks athletes' competition
results over time to identify unusual patterns that may warrant investigation
\cite{schumacher2009profiling,iljukov2017_performance_profiling,
hopker2018performance}. This approach is motivated by the observation that
performance improvement is the primary goal of doping, making competition
outcomes a natural screening signal
\cite{schumacher2009profiling,iljukov2017_performance_profiling}.

A central challenge is context sensitivity: the same race time may be expected
in one setting but anomalous in another. For example, in 100\,m sprints, altitude provides
a time advantage of approximately 0.19\,s at 2250m elevation
\cite{linthorne2016_altitude}. Additionally, a maximum legal tailwind of +2.0\,m/s
improves times by approximately 0.10 seconds \cite{linthorne1994_wind}.
Methods that treat all performances as directly comparable, ignoring wind,
altitude, and competition round, systematically inflate false positives by
flagging legitimately fast performances as anomalous
\cite{hopker2018performance,iljukov2017_performance_profiling}.

Recent work formalizes trajectory-based monitoring through longitudinal
statistical models. For example, the excess performance approach fits a career progression curve to each athlete's historical results, then measures deviation of new
performances from this expected trajectory
\cite{hopker2020_performance_profiling,hopker2024_excess_performance}. Hopker
et al. demonstrate this approach on shot putters and 100\,m sprinters, showing
that athletes with confirmed doping sanctions exhibit larger trajectory
deviations than presumed clean athletes
\cite{hopker2024_excess_performance}. This method serves as our primary
reproducible baseline.

Bayesian hierarchical models offer an alternative that explicitly represents
uncertainty and handles athletes with sparse data \cite{montagna2018_bayesian}.
These models simultaneously estimate population-level performance distributions
and individual athlete deviations. For athletes with sparse data, the model
borrows statistical strength from the population distribution, stabilizing
estimates while allowing individual trajectories to diverge from population
trends when supported by sufficient data
\cite{montagna2018_bayesian,gelman2013bayesian}. However, Bayesian methods
require repeated likelihood evaluations across thousands of MCMC sampling
iterations, and inference for tens of thousands of athletes can require
hours to days of computation \cite{gelman2013bayesian,montagna2018_bayesian},
limiting interactive screening applications.

\subsection{Operational Context and Prior Work Limitations}
Performance monitoring should function as an investigative screening tool that
prioritizes athletes for targeted biological testing or expert review, not as
an enforcement mechanism \cite{schumacher2009profiling,
iljukov2017_performance_profiling,hopker2018performance}. Performance
improvements have many legitimate explanations including training adaptations,
maturation, recovery from injury, equipment advances, and nutritional changes
\cite{hopker2018performance,iljukov2017_performance_profiling}. The World
Anti-Doping Code specifies that analytical findings (direct detection of
prohibited substances in biological samples) or Athlete Biological Passport
abnormalities are required to establish anti-doping rule violations;
performance data alone is insufficient to establish violations
\cite{wada_isti_2023,hopker2018performance}.

The Athlete Biological Passport (ABP) tracks hematological and steroidal
biomarkers over time using adaptive Bayesian models combined with expert
review panels \cite{sottas2011abp,wada_abp_og_2023}. However, the ABP operates
on laboratory measurements collected through targeted sample collection, not
on competition outcomes. Performance monitoring research positions competition
results as a complementary screening signal: unusual performance patterns can
prioritize athletes for more intensive biological testing, with the biological
evidence providing the definitive basis for sanctions
\cite{hopker2018performance,iljukov2017_performance_profiling}.

Prior work in this area faces three practical limitations. First, existing studies evaluate
a small number of detection methods (typically 1 to 3 per paper) on limited
athlete populations \cite{montagna2018_bayesian,
hopker2020_performance_profiling,ryoo2022_weightlifting}. While these studies
provide valuable methodological advances, the lack of common evaluation
protocols and datasets makes cross-method comparison difficult. Secondly,
evaluation is complicated by incomplete ground truth: publicly available
sanctions reflect only detected violations and can arrive 2 to 8 years after
the original performance due to sample storage and reanalysis policies
\cite{wada_sample_storage_2020,ita_london2012_reanalysis}. Finally, most prior
work treats performance monitoring as a research signal for post-hoc analysis
rather than as production-ready decision support \cite{hopker2018performance}. Our work addresses these gaps by providing: (1) a reproducible data pipeline
that links athlete identities across 19,000+ competitions and integrates
publicly available sanction records, enabling large-scale evaluation; (2)
systematic benchmarking of eight detection methods on 31,604 athletes under
identical evaluation protocols, with excess performance
\cite{hopker2024_excess_performance} as the primary reproducible baseline;
and (3) an integrated visual analytics system that enables interactive
screening, method comparison, and transparent case review rather than opaque
automated scoring.

\section{System Architecture and Data Pipeline}\label{system-architecture}
We present a production-ready system for performance anomaly detection that
ingests public competition data, supports multiple detection methods, and
provides investigator-facing visual analytics. The architecture prioritizes
three design goals: (1) scalability to millions of performance records via
indexed relational storage, (2) method extensibility through a standardized
detection interface enabling fair comparison, and (3) separation of concerns
between data collection, storage, analysis, and presentation to support
reproducible research and independent method evaluation.

\begin{figure}[H]
\centering
\includegraphics[width=\linewidth]{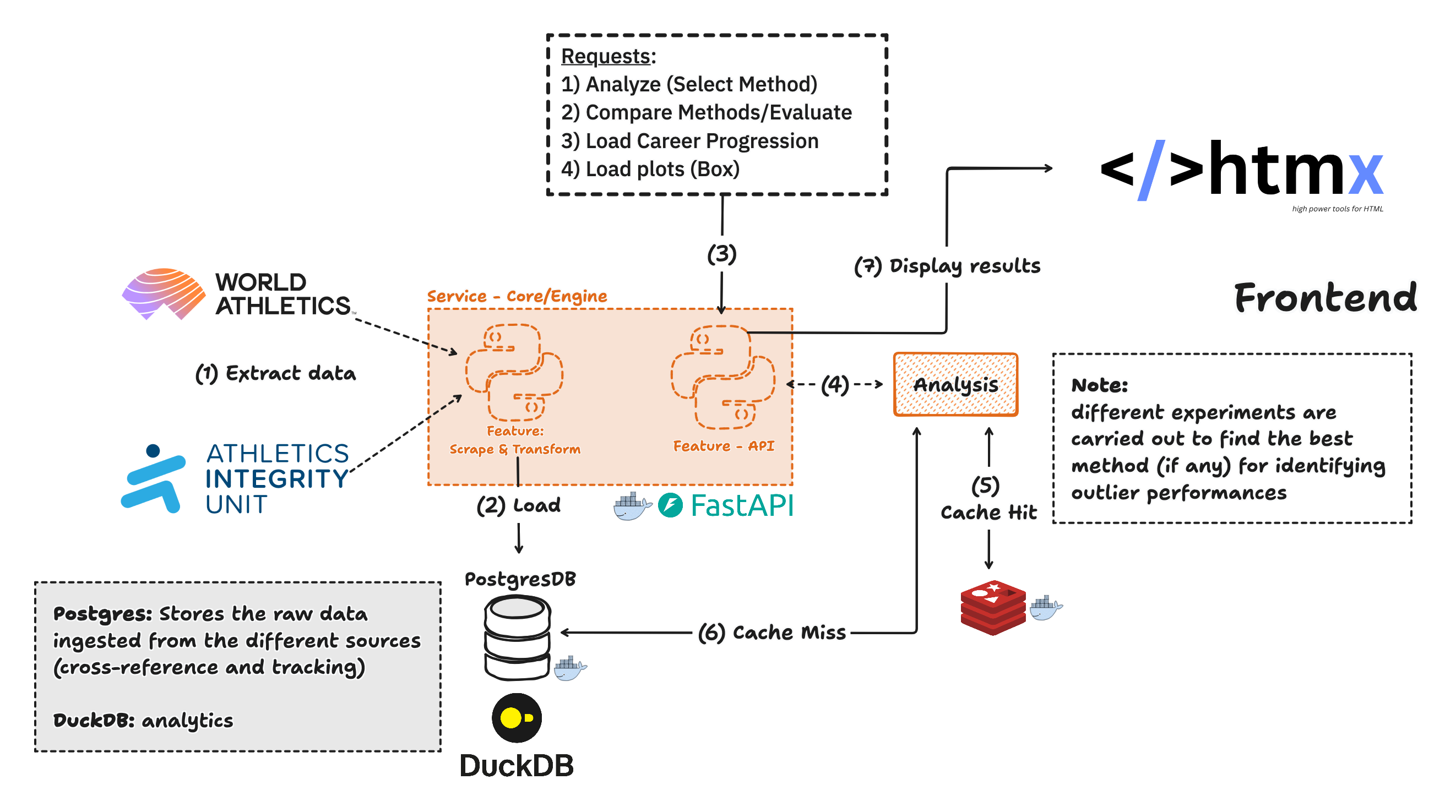}
\caption{Service-oriented architecture for performance anomaly detection.
Public athletics data flows through ingestion, storage, modular analysis
services, and interactive visualization, with clear separation between
components to support extensibility and reproducible evaluation.}
\label{fig:architecture}
\end{figure}

The system follows a service-oriented architecture with four main components
(Fig.~\ref{fig:architecture}): data acquisition and transformation, persistent
storage, anomaly detection services, and an interactive web interface. We
ingest competition results from \href{https://worldathletics.org}{World Athletics}
results pages and sanction records from the Athletics Integrity Unit's public
ineligibility database. Results are normalized by standardizing country codes,
parsing ISO timestamps, converting performance strings to seconds, and geocoding
venue locations, then loaded into PostgreSQL for durable storage and efficient
querying.

\subsection{Data Collection and Storage}
Data collection is handled by a Python ingestion service that manages
pagination and interruption recovery. Table~\ref{tab:dataset_stats} summarizes
the current dataset coverage, which enables longitudinal analysis of athlete
trajectories and evaluation against publicly available sanction records.

\begin{table}[h]
\centering
\caption{Dataset statistics for athletics performance data (2010--2025)}
\label{tab:dataset_stats}
\begin{tabular}{lr}
\toprule
\textbf{Metric} & \textbf{Count} \\
\midrule
Competitions & 19,712 \\
Individual performances & 1,583,108 \\
Unique athletes & 214,914 \\
Events covered & 103 \\
Venues & 4,251 \\
Countries represented & 221 \\
Sanctioned athletes (ground truth) & 60 \\
\bottomrule
\end{tabular}
\end{table}

\textbf{Environmental Data Handling.} Wind data differ by event: short sprints (100\,m, 100\,mh, 110mh) have 98\%+ coverage, 200\,m has 83\%, while 400\,m events have 0\% (not measured for longer races). Sprint events use the \texttt{wind\_legal\_results} database view which filters illegal wind (>+2.0\,m/s per World Athletics rules) while retaining missing values. Missing wind values are imputed with 0 for ML-based methods, which conflates neutral conditions (0.0 m/s) with unknown conditions. Altitude is not adjusted in the current version. All rounds (heats, semifinals, finals) are retained in analysis.

PostgreSQL serves as the primary persistent store with B-tree indexing on
\texttt{athlete\_id} and \texttt{competition\_date} for transactional queries.
The schema comprises three core tables with foreign key relationships:
\texttt{raw\_results} stores individual performance records (athlete
identifier, time, rank, wind speed, reaction time, round context) linked to
\texttt{crawled\_competitions} via \texttt{competition\_id} (primary key);
\texttt{crawled\_competitions} stores competition metadata (date, location,
venue, event type, environmental conditions); \texttt{ineligible\_persons}
stores sanction records from the Athletics Integrity Unit, linked to athletes
via \texttt{athlete\_id} for ground truth evaluation. Analytical queries, including
athlete trajectory aggregation, cross-athlete comparisons, and method
evaluation, are executed through DuckDB, which provides columnar storage and
vectorized query processing for sub-second analytical performance over 1M+
records. DuckDB attaches to PostgreSQL tables via the \texttt{postgres\_scanner}
extension, enabling zero-copy access to persistent data while maintaining
separation between transactional and analytical workloads. Redis provides
caching and asynchronous task queuing for long-running analyses. Screening queries over 381,447 100\,m performances return in under 0.5\,s with Redis caching enabled.

\subsection{Anomaly Detection Service}
Anomaly detection is implemented through a unified analysis service that
exposes multiple detection methods via a standardized interface. Each method
implements the contract: \texttt{detect(performance\_data) → DetectionResult},
where \texttt{performance\_data} is a DataFrame containing athlete histories
with performance times, dates, wind speeds, competition metadata, and venue
information. The returned \texttt{DetectionResult} structure contains outlier
classifications, anomaly scores, method-specific explanations, and performance
timing metrics. This standardization enables fair comparison: all methods
operate on identical athlete histories and contextual features under identical
evaluation protocols.

\subsection{Investigator Interface}
\begin{figure}[htb]
\centering
\includegraphics[width=\columnwidth]{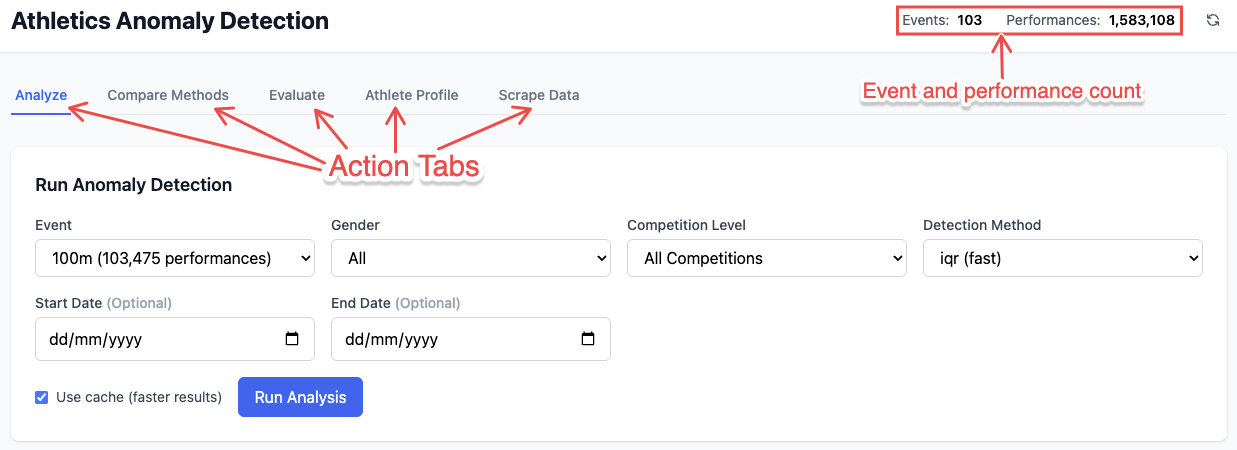}
\caption{Investigator workflow: coarse-to-fine filtering (event $\rightarrow$ gender $\rightarrow$ date $\rightarrow$ method) for screening large athlete groups. Displays flagged athletes + method-agreement indicators to support consensus-based triage and prioritization.}
\label{fig:screening_workflow}
\end{figure}

The web application, built with HTMX for server-driven interactivity, supports two primary workflows. First, \textit{screening}
identifies candidates for review: investigators filter by event (e.g., 100\,m
men) and time window (e.g., 2020--2025), then select a detection method to
flag athletes with anomalous performance patterns (Fig.~\ref{fig:screening_workflow}). Second, \textit{case review}
examines individual athlete trajectories: the interface displays performance
progression over time, overlays anomaly scores from selected methods,
highlights flagged performances with method-specific explanations, and provides
direct links to source competition results for verification. 

\begin{figure}[htb]
\centering
\includegraphics[width=\columnwidth]{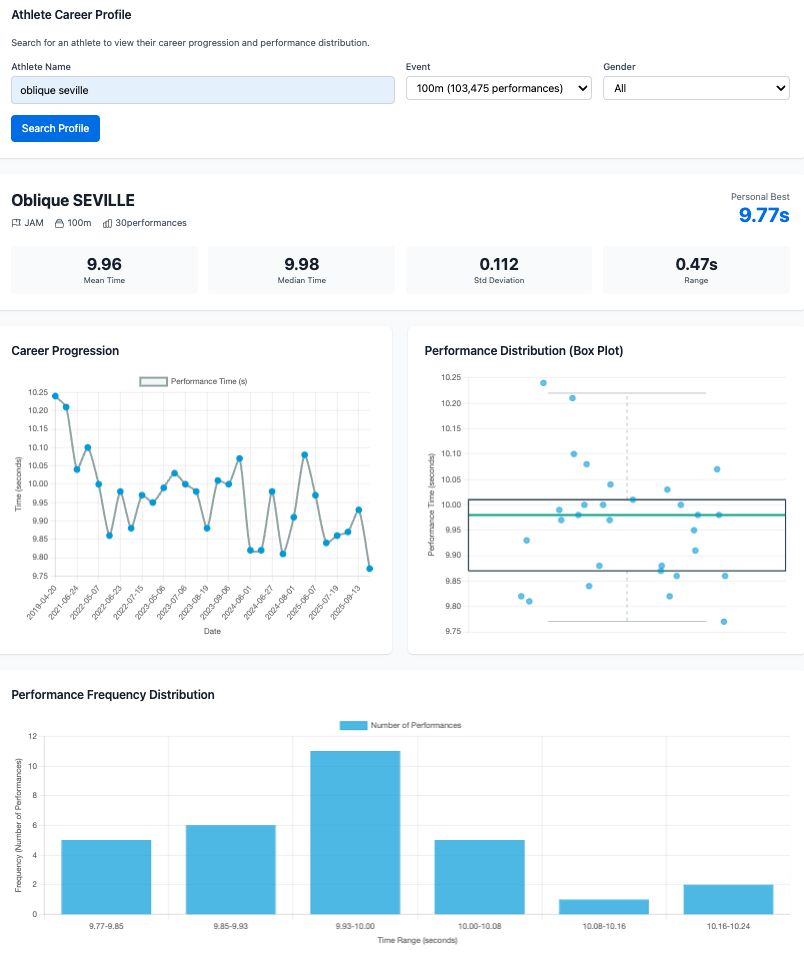}
\caption{An athlete's distribution plots including box plots and histograms. These show the full range of results, revealing whether performances are mostly steady (tight spread) or include rare extremes (long tails), which helps when judging threshold-based methods.}
\label{fig:dash_distribution}
\end{figure}

Distributional views (Fig.~\ref{fig:dash_distribution})
expose whether an athlete has occasional extreme performances
or consistent results, aiding interpretation of threshold-based
detectors. This transparency
enables investigators to understand both individual method decisions and
cross-method consensus, supporting expert review rather than automated enforcement. A live demo is available at \url{https://athletics-performance.mblessed.space}.

\section{Detection Methods}\label{detection-methods}
We implement eight detection methods spanning the main approaches in sports
performance anomaly detection: statistical rules
\cite{grubbs1969procedures,hoaglin1983robust}, machine learning classifiers
\cite{liu2008isolation,chen2016xgboost}, temporal trajectory models
\cite{hopker2024_excess_performance}, Bayesian hierarchical inference
\cite{montagna2018_bayesian}, and multivariate dependence modeling
\cite{nelsen2006copulas}. All methods and their computational complexity are summarized in Table~\ref{tab:methods_summary}.

\begin{table}[htb]
\centering
\caption{Summary of Implemented Detection Methods}
\label{tab:methods_summary}

\begin{minipage}{\columnwidth}
\centering
\scriptsize
\setlength{\tabcolsep}{3.5pt}

\begin{tabular}{@{}c l p{4.0cm} l@{}}
\toprule
\textbf{CT} & \textbf{Method} & \textbf{Key Idea} & \textbf{Complexity} \\
\midrule
ST & Z-Score & Performances $>3\sigma$ from mean & $O(n)$ \\
ST & IQR Method & Tukey fences (1.5×IQR) & $O(n \log n)$ \\
ST & Robust Z (MAD) & Median deviation threshold & $O(n \log n)$ \\
\midrule
ML & Isolation Forest & Fewer splits to isolate points & $O(n \log n)$ \\
ML & XGBoost & Residuals $>95\%$ & $O(n \log n)$ \\
\midrule
TM & Excess Perf. & Career trajectory deviation & $O(n)$ \\
\midrule
BS & Hierarchical & Posterior pred. $p<0.05$ & $O(n \cdot \text{MCMC})$ \\
\midrule
MV & Gauss. Copula & Joint density bottom $5\%$ & $O(n^2)$ \\
\bottomrule
\end{tabular}

\vspace{2mm}
\footnotesize
CT: Category. ST=Statistical, ML=Machine Learning, TM=Temporal, BS=Bayesian, MV=Multivariate.

\end{minipage}
\end{table}

\vspace{-3mm}

\subsection{Statistical Methods}
Statistical methods flag performances deviating substantially from an athlete's
historical distribution using univariate outlier rules.

\textbf{Z-Score \cite{grubbs1969procedures}:} 
Flags performances that deviate more than $k$ standard deviations from the athlete's career mean:
\begin{equation}
z_i = \frac{x_i - \mu}{\sigma}, 
\quad \text{flag if } |z_i| > 3.0
\end{equation}
where $x_i$ is performance time, $\mu$ is the career mean, and $\sigma$ is the career standard deviation. 
Under normality, this flags approximately 0.3\% of observations.

\textbf{Robust Z-Score (Median + MAD) \cite{hoaglin1983robust}:} 
Uses median-based statistics that are less sensitive to extreme values:
\begin{equation}
z_i^* = \frac{0.6745 \cdot (x_i - \mathrm{median}(x))}{\mathrm{MAD}(x)}, 
\quad \text{flag if } |z_i^*| > 3.5
\end{equation}
where
\begin{equation}
\mathrm{MAD}(x) = \mathrm{median}(|x_i - \mathrm{median}(x)|)
\end{equation}
The constant 0.6745 scales MAD to match the standard deviation under normality.
This approach is more robust because the median and MAD are less influenced by extreme performances.

\textbf{IQR Method \cite{hoaglin1983robust}:} Applies Tukey's fence rule,
flagging values outside
\begin{equation}
[Q_1 - 1.5 \cdot \mathrm{IQR},\; Q_3 + 1.5 \cdot \mathrm{IQR}]
\end{equation}
where IQR = $Q_3 - Q_1$. The factor 1.5 is the standard Tukey fence multiplier
\cite{hoaglin1983robust}.

\subsection{Machine Learning Methods}
Machine learning methods learn decision boundaries or density models from
performance histories, enabling anomaly detection without parametric
assumptions. We use default scikit-learn and XGBoost hyperparameters throughout 
(detailed in Table~\ref{tab:hyperparameters}). Tuning on sanctioned athletes 
would introduce look-ahead bias, as sanction labels must remain reserved for 
evaluation.

\textbf{Isolation Forest \cite{liu2008isolation}:} Builds an ensemble of
random decision trees; anomalies require fewer splits to isolate because they
lie in sparse feature space regions. Anomaly score:
\begin{equation}
s(x, n) = 2^{-\frac{E(h(x))}{c(n)}}
\end{equation}
where $h(x)$ is average path length required to isolate $x$, and $c(n)$
normalizes by expected path length in a binary tree of size $n$. Scores near
1 indicate anomalies. We engineer features capturing performance (time),
context (wind speed, competition level), and recent form (5-race moving
average). Training: 100 trees, contamination parameter 0.1.

\textbf{XGBoost Residual \cite{chen2016xgboost}:} Trains a gradient boosting
regressor to predict performance time from contextual features (wind, altitude,
round, competition level). We flag residuals exceeding the 95th percentile:
\begin{equation}
r_i = |y_i - \hat{y}_i|, \quad \text{flag if } r_i > Q_{95}
\end{equation}
The 95th percentile corresponds approximately to $2\sigma$ under normality
while remaining robust to heavy tails. Hyperparameters: 100 trees, learning
rate 0.1, max depth 3.

\subsection{Temporal Methods}
Temporal methods model athlete performance progression over time, flagging
deviations from expected trajectories.

\textbf{Excess Performance \cite{hopker2024_excess_performance}:} Compares each performance to the athlete's personal baseline -- career mean $\mu_i$ and standard deviation $\sigma_i$ for athletes with $\geq$3 performances -- and flags substantial deviations:

\begin{equation}
\text{EP}_i = \frac{x_i - \mu_i}{\sigma_i}, \quad \text{flag if } \text{EP}_i < -2.5
\end{equation}
This per-athlete normalization accounts for inter-athlete variability but uses a static baseline, ignoring career progression. Unlike trajectory-based smoothing curve methods \cite{hopker2024_excess_performance}, it prioritizes efficiency and robustness for sparse data (median 6–8 competitions per athlete).

\subsection{Bayesian Methods}
Bayesian hierarchical models estimate population-level performance
distributions and athlete-specific deviations simultaneously, enabling
uncertainty quantification and borrowing of statistical strength across
athletes \cite{montagna2018_bayesian}.

\textbf{Hierarchical Linear Model \cite{gelman2013bayesian}:} Models athlete
performance trajectories with random intercepts and slopes:
\begin{align}
y_{ij} &\sim \mathcal{N}(\alpha_i + \beta_i \cdot t_j, \sigma^2) \\
\alpha_i &\sim \mathcal{N}(\mu_\alpha, \tau_\alpha^2) \\
\beta_i &\sim \mathcal{N}(\mu_\beta, \tau_\beta^2)
\end{align}
where $y_{ij}$ is the $j$-th performance of athlete $i$, $t_j$ is time,
$\alpha_i$ and $\beta_i$ are athlete-specific intercept and slope, and
hyperparameters capture population-level trends. We use weakly informative
priors: $\mu_\alpha \sim \mathcal{N}(11, 1)$, $\mu_\beta \sim \mathcal{N}(0,
0.1)$, $\tau_\alpha, \tau_\beta \sim \text{HalfNormal}(1)$, centered on
typical 100\,m times and improvement rates. Inference via No-U-Turn Sampler
(NUTS) with 500 draws per chain (4 chains, 200 warmup). We flag
performances with posterior predictive p-value $< 0.05$, corresponding
approximately to $2\sigma$ under normality while accounting for model
uncertainty.

\subsection{Multivariate Methods}
Multivariate methods model joint distributions across multiple performance
dimensions, detecting anomalies in dependence structure rather than marginal
distributions alone.

\textbf{Gaussian Copula \cite{nelsen2006copulas}:} Models the joint
distribution of performance time, wind speed, and reaction time using a copula
separating marginal distributions from dependence structure:
\begin{equation}
C(u_1, \ldots, u_d) = \Phi_R(\Phi^{-1}(u_1), \ldots, \Phi^{-1}(u_d))
\end{equation}
where $\Phi$ is the standard normal CDF, $R$ is the correlation matrix, and
$u_i$ are marginal cumulative probabilities. These three features have
complete data coverage ($>$80\% of performances) and capture outcome (time),
context (wind), and execution (reaction time). We flag points in the bottom 5\% of copula density — combinations that are unusual jointly but would not be detected by univariate methods alone.

\section{Evaluation}\label{evaluation}
We evaluate detection methods on large-scale sprint data to assess which
approaches best identify sanctioned athletes as candidates for further review,
measuring their ability to surface athletes with confirmed anti-doping rule
violations while quantifying precision-recall tradeoffs.

\subsection{Evaluation Setting and Ground Truth}
\textbf{Ground Truth.} We use publicly available ineligibility records from the Athletics Integrity Unit. These sanctions constitute incomplete labeling in two ways: undetected dopers reduce measured recall (true positives go uncounted), and flagged athletes who are doping but unsanctioned reduce precision (counted as false positives). Therefore, reported metrics (F1=0.016, Precision=0.009) represent conservative lower bounds. Many flagged athletes may represent undetected cases rather than false positives.

\textbf{Evaluation Protocol.} Methods are evaluated at the athlete level: a
true positive is a sanctioned athlete flagged in at least one performance.
This reflects operational screening requirements: a system should detect
suspicious athletes even if not all their performances are anomalous. We
report precision (fraction of flagged athletes who are sanctioned), recall
(fraction of sanctioned athletes flagged), and F1 score.

\subsection{Primary Evaluation Slice: 100\,m Sprint}
We focus on 100\,m sprint results because: (1) it has the most complete data
coverage, (2) contextual effects (wind, altitude) are well-characterized
\cite{linthorne2016_altitude,linthorne1994_wind}, enabling fair comparison of
context-aware methods, and (3) the 100\,m represents the highest-profile event
in athletics. The 100\,m evaluation slice contains 31,604 athletes with 381,447
recorded performances, of which 25 athletes have confirmed sanctions (0.08\%
positive class rate). This severe class imbalance is typical of anti-doping
screening contexts.

\subsection{Detection Results}

\begin{figure}[htb]
\centering
\includegraphics[width=\columnwidth]{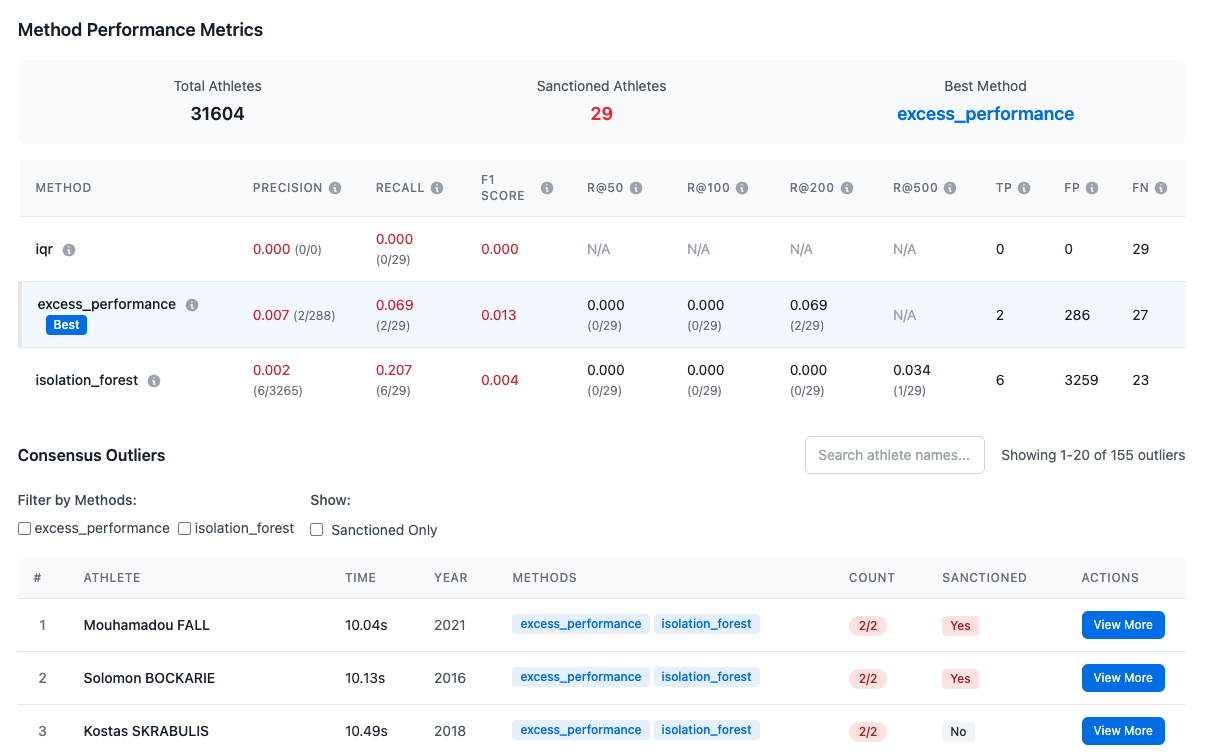}
\caption{Consensus outliers interface. Athletes flagged by multiple methods are
highlighted, with filters for method selection and sanction status. Athletes
flagged by 3+ independent methods are prioritized for review, reducing false
positives through cross-method agreement.}
\label{fig:dash_consensus}
\end{figure}

Athlete-level results are shown in Table~\ref{tab:method_performance}. Excess Performance has the best F1 score (0.016). Bayesian Hierarchical reaches an F1 of 0.011. Isolation Forest has the highest recall (0.240) but low precision (0.003), flagging 1,854 athletes to find 6 sanctioned cases. Five methods find no sanctioned athletes. Precision@200 for Excess Performance is 0.010 (2 of the top 200 flagged are sanctioned). Athletes flagged by multiple methods are ranked higher for review (Fig.~\ref{fig:dash_consensus}), which helps cut false positives when review time is limited.

\begin{table}[htb]
\centering
\footnotesize
\caption{Detection Method Performance on 100\,m Sprint}
\label{tab:method_performance}
\begin{tabular}{@{}lcccccc@{}}
\toprule
\textbf{Method} & \textbf{P} & \textbf{R} & \textbf{F1} & \textbf{P@200} & \textbf{TP} & \textbf{FP} \\
\midrule
Excess Perf.$^{*}$ & .009 & .080 & .016 & .010 & 2 & 224 \\
Bayesian Hier. & .006 & .160 & .011 & --- & 4 & 707 \\
Isolation For. & .003 & .240 & .006 & .000 & 6 & 1848 \\
IQR$^{\dagger}$ & .000 & .000 & .000 & --- & 0 & 5 \\
Copula$^{\dagger}$ & .000 & .000 & .000 & --- & 0 & 16 \\
XGBoost$^{\dagger}$ & .000 & .000 & .000 & --- & 0 & 53 \\
MAD$^{\ddagger}$ & .000 & .000 & .000 & --- & 0 & 0 \\
Z-Score$^{\ddagger}$ & .000 & .000 & .000 & --- & 0 & 0 \\
\bottomrule
\end{tabular}
\vspace{1mm}
\begin{flushleft}
\scriptsize
31,604 athletes, 25 sanctioned. P=Precision, R=Recall, P@200=Precision at rank 200.\\
$^{*}$Baseline. P@200=0.010: 2/200 sanctioned (1\% hit rate).\\
$^{\dagger}$Flags athletes, detects none. $^{\ddagger}$No flags.
\end{flushleft}
\end{table}

\textbf{Method Failures and Tradeoffs.} Five methods achieve zero F1 for different reasons. Simple statistical thresholds (Z-Score, MAD) produce no flags because typical within-athlete performance variability falls well within their threshold bounds, leaving no single performance far enough from the athlete's career mean to trigger detection. IQR, Gaussian Copula, and XGBoost Residual flag 5–54 athletes but detect no sanctioned individuals, capturing unusualness unrelated to violations. The three methods with non-zero detection show tradeoffs: Excess Performance is conservative (226 flagged, 2 detected), suitable for limited resources. Isolation Forest is aggressive (1,854 flagged, 6 detected), giving 3× higher recall but 3× lower precision. Bayesian Hierarchical is intermediate (711 flagged, 4 detected).

\subsection{Generality Across Sprint Events}
We evaluate the top-performing methods on 200\,m and 400\,m sprints using the same athlete-level protocol as the 100\,m benchmark (Section~\ref{evaluation}). Publicly available sanctions serve as partial ground truth. The 200\,m slice includes 19,928 athletes (25 sanctioned), while the 400\,m slice includes 20,491 athletes (26 sanctioned). As shown in Table~\ref{tab:multi_event_summary}, the best-performing method varies by event: Excess Performance achieves the highest F1 on 100\,m (0.016), IQR on 200\,m (0.095), and Gaussian Copula on 400\,m (0.050). Whether these event-specific differences generalize requires further investigation.

\begin{table}[htb]
\centering
\caption{Cross-Event Method Comparison (Sprint Events)}
\label{tab:multi_event_summary}
\begin{tabular}{@{}lcccc@{}}
\toprule
\textbf{Event} & \textbf{Athletes} & \textbf{Sanctioned} & \textbf{Best F1} & \textbf{Method} \\
\midrule
100\,m & 31,604 & 25 & 0.016 & Excess Performance \\
200\,m & 19,928 & 25 & 0.095 & IQR Method \\
400\,m & 20,491 & 26 & 0.050 & Gaussian Copula \\
\bottomrule
\end{tabular}
\begin{flushleft}
\footnotesize
Best method varies despite similar sanction rates (0.08\%--0.13\%).
\end{flushleft}
\end{table}


\section{Discussion}\label{discussion}

We present an online visual analytics system for examining anomalous athletics performances, supporting cross-method comparison, career trajectory visualization, and consensus-based screening.

\textbf{Screening Design and Low F1.} F1 scores are low (0.6–1.6\%) because this is a screening system, not a binary classifier. It narrows 31,604 athletes to about 163 consensus outliers  (athletes flagged by $\geq$2 methods; 99.5\% reduction), achieving 10–18× better detection than random sampling (0.08\%). Investigators combine algorithmic flags with biological passport data, whereabouts, and coaching networks to prioritize testing. False positives undergo review, not costly biological testing (\$800 per sample). The system supports decision-making, not automated enforcement.

\textbf{Event-Specific Performance Variation.} Differences across events (Table~\ref{tab:multi_event_summary}) reflect physiological factors. Sprints (100\,m, 200\,m) have tight performance spreads (~0.12s), favoring density-based methods like Isolation Forest. No single method dominates all events, validating our multi-method ensemble.

The three methods detecting sanctioned athletes show tradeoffs. Excess Performance is conservative, Isolation Forest flags more athletes at lower precision, and Bayesian Hierarchical is intermediate. Detection rates match biological testing (0.80\% positive rate \cite{wada_testing_figures_2023}) and confirm that performance screening prioritizes rather than replaces targeted testing. Future work could combine methods to improve precision at a fixed recall.

\textbf{Limitations.} Sparse ground truth (25 of 31,604 athletes in 100\,m; 60 total across sprint events) makes metrics conservative since some flagged athletes may be true but undetected cases. Wind data vary by event (as detailed in Section~\ref{system-architecture}), altitude effects are unadjusted, which may inflate flags for high-elevation competitions, and public sanctions may appear 2 to 8 years after the performance \cite{wada_sample_storage_2020,ita_london2012_reanalysis}. The system’s main value is decision support: reducing the review scope by 99.5\% and generating investigative leads for expert review using only publicly available data. It must not be used to accuse athletes without corroborating biological evidence.

\section{Conclusion}\label{conclusion}
This work demonstrates that large-scale performance data can support
screening-oriented decision support despite incomplete ground truth and missing
environmental data. We benchmark eight detection methods on 31,604 athletes,
quantifying precision-recall tradeoffs across statistical, machine learning,
and Bayesian approaches. Excess Performance achieves best F1 (0.016), while
Isolation Forest achieves higher recall (0.240) at lower precision (0.003).
The evaluation protocol generalizes across 100\,m, 200\,m, and 400\,m sprints,
demonstrating event-agnostic design.

\section*{Acknowledgments}
This publication was developed as part of the Center for Inclusive Digital Transformation of Africa (CIDTA) and the Afretec Network, which is managed by Carnegie Mellon University Africa and receives financial support from the Mastercard Foundation. The views expressed in this document are solely those of the authors and do not necessarily reflect those of the Carnegie Mellon University or the Mastercard Foundation. We thank World Athletics and the Athletics Integrity Unit for public access to results and sanctions data.

\bibliographystyle{IEEEtran}
\bibliography{sds_references}

\appendices

\section{Hyperparameter Configuration}
\label{app:hyperparameters}

\begin{table}[H]
\centering
\caption{ML-Based Detection Methods Hyperparameters}
\label{tab:hyperparameters}
\begin{tabular}{@{}llll@{}}
\toprule
\textbf{Method} & \textbf{Parameter} & \textbf{Value} & \textbf{Justification} \\
\midrule
\multirow{3}{*}{Isolation Forest} & contamination & 0.1 & Domain prior: $\sim$10\% outliers \\
 & n\_estimators & 100 & sklearn default \\
 & random\_state & 42 & Reproducibility \\
\midrule
\multirow{3}{*}{XGBoost} & n\_estimators & 100 & Standard for regression \\
 & max\_depth & 3 & Prevent overfitting \\
 & learning\_rate & 0.1 & sklearn default \\
\midrule
\multirow{2}{*}{Bayesian} & MCMC draws & 500 & Posterior sampling \\
 & tune (burn-in) & 200 & Chain convergence \\
\bottomrule
\end{tabular}
\end{table}

\end{document}